\documentclass[final,5p,times,twocolumn]{elsarticle}

\usepackage{amsmath}
\usepackage{mathtools}
\usepackage{xcolor}
\usepackage{enumitem}
\usepackage{comment}
\usepackage{graphicx}
\usepackage{multirow}
\usepackage{array}
\usepackage{caption}
\usepackage{fontawesome5}
\usepackage{hyperref}
\usepackage{wrapfig}
\usepackage{longtable}
\usepackage{times}
\usepackage{latexsym}
\usepackage{geometry}
\usepackage{array}
\usepackage{latexsym}
\usepackage{color, colortbl}
\usepackage{booktabs}

\usepackage{algorithm} 
\usepackage{algpseudocode}

\usepackage{amssymb}

\begin{document}

\begin{frontmatter}



\title{Semantic-Driven Topic Modeling for Analyzing Creativity in Virtual Brainstorming}

\author[1]{Melkamu Abay Mersha\corref{cor1}}

\author[1]{Jugal Kalita}

\affiliation[1]{organization={College of Engineering and Applied Science, University of Colorado Colorado Springs},
            addressline={Colorado Springs},
            postcode={80918},
            state={CO},
            country={USA}}



\cortext[cor1]{Corresponding author.}  

\cortext[email]{\textit{E-mail addresses:} 
\href{mailto:melkamu.mersha@uccs.edu}{mmersha@uccs.edu} (M.A. Mersha*), 
\href{mailto:jkalita@uccs.edu}{jkalita@uccs.edu} (J. Kalita).}

\begin{abstract}
Virtual brainstorming sessions have become a central component of collaborative problem solving, yet the large volume and uneven distribution of ideas often make it difficult to extract valuable insights efficiently. Manual coding of ideas is time-consuming and subjective, underscoring the need for automated approaches to support the evaluation of group creativity. In this study, we propose a semantic-driven topic modeling framework that integrates four modular components: transformer-based embeddings (Sentence-BERT), dimensionality reduction (UMAP), clustering (HDBSCAN), and topic extraction with refinement. The framework captures semantic similarity at the sentence level, enabling the discovery of coherent themes from brainstorming transcripts while filtering noise and identifying outliers. We evaluate our approach on structured Zoom brainstorming sessions involving student groups tasked with improving their university. Results demonstrate that our model achieves higher topic coherence compared to established methods such as LDA, ETM, and BERTopic, with an average coherence score of 0.687 (\textit{CV}), outperforming baselines by a significant margin. Beyond improved performance, the model provides interpretable insights into the depth and diversity of topics explored, supporting both convergent and divergent dimensions of group creativity. This work highlights the potential of embedding-based topic modeling for analyzing collaborative ideation and contributes an efficient and scalable framework for studying creativity in synchronous virtual meetings.

\end{abstract}

\begin{keyword}
Semantic-driven topic modeling; Sentence-BERT; UMAP; HDBSCAN; Virtual brainstorming; Collaborative creativity; Topic coherence

\end{keyword}

\end{frontmatter}

\section{Introduction}

Digital communication has become central to modern collaboration, with virtual meetings and conversational agents such as chatbots increasingly shaping how teams interact across geographical and cultural boundaries \cite{uthirapathy2023topic}. In organizational settings, much of this interaction occurs through text, where individuals exchange ideas to solve problems and reach decisions. However, text-based discussions often produce a large mixture of task-relevant and off-task content \cite{tonja2023first, ziegler2000idea}. As a result, participants may find it difficult to identify which ideas are most valuable for decision-making \cite{rietzschel2010selection}. Furthermore, brainstorming in distributed teams is not always uniform: some topics are examined in great detail while others receive only superficial attention \cite{rietzschel2007relative}. These imbalances complicate the analysis of group discussions and motivate the development of automated techniques that can complement or replace manual annotation, which is costly and time-consuming.

Earlier studies introduced software tools to analyze aspects such as word count, information sharing, and non-verbal cues in group brainstorming contexts \cite{kim2008meeting,huber2019automatically}. Building on these directions, our work focuses on measuring group creativity through the identification of latent themes and assessing how deeply groups explore those themes. Such measures are critical, since creative performance depends on both convergent and divergent thinking within group interactions. Recent advances in large language models (LLMs) have highlighted the potential of automated systems to capture human-like reasoning and language understanding \cite{mersha2024ethio, yigezu2024ethio}. Motivated by these developments, we also investigate whether such models can perform topic identification tasks as effectively as manual or traditional machine learning approaches.

Machine learning methods, particularly topic modeling, provide a structured way to uncover interpretable patterns from large volumes of text. Topic models can organize documents into latent themes and reveal semantic relationships across ideas, making them useful for data exploration, summarization, information retrieval, recommendation, and other applications \cite{blei2012probabilistic,rodriguez2020computational,joshi2023deepsumm,albalawi2020using,punziano2023digital,mustak2021artificial,kim2019insider,sbalchiero2020topic,trenk2024text}. Classical techniques such as Latent Dirichlet Allocation (LDA) \cite{blei2003latent}, Non-Negative Matrix Factorization (NMF) \cite{fevotte2011algorithms}, and Latent Semantic Analysis (LSA) \cite{hofmann1999probabilistic} remain widely used but generally rely on statistical co-occurrence patterns without explicitly accounting for semantic similarity between words or sentences.

Capturing semantic relatedness is particularly important in brainstorming contexts, where creativity often emerges from linking distant concepts or exploring topics at varying depths. Research on the creative process shows that the novelty of an idea frequently increases as the semantic distance between concepts grows, and that deeper exploration within a single topic can lead to the discovery of unconventional or original insights \cite{nijstad2006group,paulus2007toward}. Measuring these semantic distances therefore provides a useful proxy for the level of divergent thinking in a group. For example, Huber et al. \cite{huber2019automatically} employed topic modeling to study group creativity in Slack conversations, but their analysis was limited to publicly accessible asynchronous messages. Extending such approaches to synchronous collaboration is critical for understanding how real-time interactions influence both productivity and creativity.

In our earlier work, Mersha et al.~\cite{mersha2024semantic} proposed a semantic-driven topic modeling framework that leverages transformer-based embeddings, dimensionality reduction, and clustering to identify coherent themes from group brainstorming transcripts. That study demonstrated the advantages of using contextual embeddings for extracting meaningful topics compared to classical probabilistic approaches. Building on this foundation, the present study applies and extends that framework to structured Zoom brainstorming sessions, where small groups were tasked with addressing a specific problem. Our primary objectives are to categorize generated ideas, identify salient topics, and analyze the depth and diversity of topic exploration. 

The proposed model integrates four main components---embedding, dimensionality reduction, clustering, and topic extraction---each designed to exploit the contextual representations offered by transformer models. By combining these components, the framework yields improved topic discovery performance while providing interpretable insights into group creativity. Our study makes the following contributions.
\begin{itemize}
    \item Proposed a semantic-driven topic modeling framework for analyzing group creativity in brainstorming sessions.  
    \item Achieved higher topic coherence and interpretability compared to traditional baseline methods.  
    \item Provided novel insights into convergent and divergent thinking patterns in collaborative group discussions.  
\end{itemize}

The paper is organized as follows: Section 2 reviews related work, Section 3 presents the proposed framework, Section 4 describes the experiments and results, Section 5 presents future work, and Section 6 concludes with key findings and future directions.

\section{Related Works}

Over the past several decades, topic modeling has become an essential method for analyzing large-scale textual data drawn from diverse sources such as chat logs, video transcripts, social media posts, and customer feedback. Within the context of collaborative ideation, applying topic modeling to categorize brainstormed ideas helps move beyond mere idea generation toward structured problem solving. Effective categorization can reduce redundancy, highlight essential contributions, and support both convergent and divergent thinking processes that underlie creative outcomes (De Vreede et al., 2022). Studies have shown that brainstorming groups often converge strongly on some topics while only superficially addressing others, and similar dual patterns are observed among design engineers who balance divergent exploration with convergent solution-finding \cite{ferguson2022communication}. Consequently, the systematic identification of topics has emerged as a critical measure of the quality of collaborative brainstorming.

Traditional approaches to assessing convergence in brainstorming relied heavily on human raters, who categorized and scored ideas generated by groups \cite{nijstad2002cognitive,nijstad1999persistence}. Although insightful, manual judgments can introduce subjectivity and scale poorly with large datasets. To address these limitations, computational techniques such as topic modeling have been proposed as automated alternatives.

Current topic modeling methods can broadly be grouped into two families: probabilistic models and embedding-based models. Probabilistic approaches—including Latent Dirichlet Allocation (LDA) \cite{blei2003latent}, Non-Negative Matrix Factorization (NMF) \cite{fevotte2011algorithms}, and Latent Semantic Analysis (LSA) \cite{hofmann1999probabilistic}—rely on co-occurrence statistics to infer latent structure in text. More recent embedding-based approaches, by contrast, exploit dense vector representations from pre-trained language models, allowing them to capture semantic similarity between words and sentences. Each paradigm offers distinct advantages and challenges \cite{koh2021topic}, and hybrid methods that integrate embeddings with probabilistic foundations have produced promising results \cite{agarwal2021comparative,qiang2017topic}.

A number of studies demonstrate the application of these methods in real-world conversational data. For instance, Chin et al. \cite{chin2023user} applied LDA to investigate user-chatbot conversations during the COVID-19 pandemic, uncovering latent topics, sentiment trends, and user attitudes toward public health issues. Similarly, Koh and Fienup \cite{koh2021topic} employed LDA to analyze library chat transcripts, showing how topic models can help improve user services. A limitation of these probabilistic approaches, however, is their inability to fully incorporate contextual information, which often leads to less coherent topics.

Embedding-based topic models have advanced the field by integrating neural embeddings, resulting in more semantically meaningful and coherent representations \cite{zhao2021topic,terragni2021octis,alshami2024smart}. For example, Bianchi et al. \cite{bianchi2020cross} and Dieng et al. \cite{dieng2020topic} showed that contextual embeddings capture fine-grained semantic relations, leading to improved topic coherence. Other works proposed clustering-based techniques that group pre-trained embeddings to accelerate topic extraction and enhance interpretability \cite{kim2019insider,tang2023research}. A hybrid model, BERTopic, combines TF–IDF with BERT embeddings to cluster semantically similar sentences and extract interpretable topics, yielding strong performance across different textual domains \cite{grootendorst2022bertopic}.

Building on these advances, our earlier study \cite{mersha2024semantic} introduced a semantic-driven framework that integrates Sentence-BERT embeddings, dimensionality reduction, and clustering for topic discovery in virtual meeting data. That work demonstrated that transformer-based embeddings provide clear advantages over classical probabilistic models when applied to group brainstorming transcripts. The present research extends this line of inquiry by applying an embedding-based model to structured brainstorming sessions, aiming to extract coherent and interpretable topics that shed light on group creativity.

\section{Methodology}
To analyze the brainstorming data, we designed a modular framework that processes all generated ideas through four sequential stages: embedding, dimensionality reduction, clustering, and topic extraction. In the first stage, each idea or sentence is transformed into a dense vector using a pre-trained transformer-based language model, which captures contextual semantics and provides a basis for measuring similarity. Since these embeddings are high-dimensional, a dimensionality reduction technique is applied in the second stage to project them into a lower-dimensional space. This step improves computational efficiency and enhances the separability of clusters. In the third stage, the reduced embeddings are grouped using a density-based clustering algorithm that identifies semantically coherent clusters of related ideas while filtering out noise or irrelevant sentences. Finally, in the topic extraction stage, each cluster is analyzed to identify the most representative words that capture its meaning. The top-$k$ words with the highest semantic relevance are then selected to define the underlying topics. 

\begin{figure}[h] 
\centering
{\includegraphics [width=0.3\textwidth]{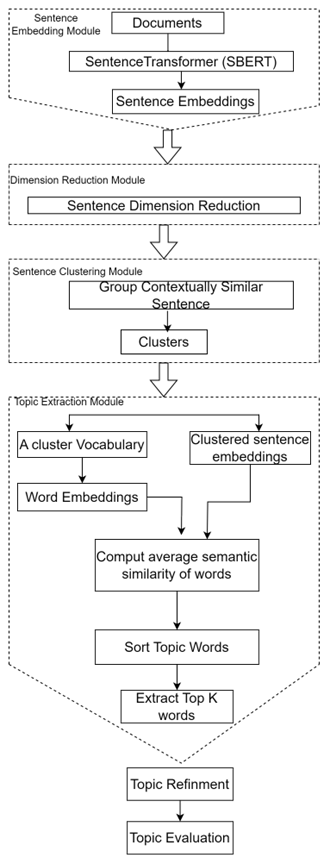}}
\caption{The proposed model architecture}
\label{fig:model}
\end{figure}

The overall architecture of the framework is illustrated in Figure~\ref{fig:model}, showing the flow from raw brainstorming transcripts to final extracted topics. By combining contextual embeddings with clustering and topic word selection, the proposed framework ensures that semantically related ideas are grouped together and summarized with interpretable labels. This enables a structured analysis of the depth and diversity of topics explored during brainstorming sessions.

\subsection{Document Embedding}

Ideas expressed during brainstorming can take many forms, including words, short phrases, and complete sentences. In this study, we treat each idea—regardless of its length—as a unit of analysis, and for simplicity we use the terms ``idea'' and ``sentence'' interchangeably. A collection of such ideas is regarded as a document. The first step of our framework involves converting these ideas into numerical representations that capture their semantic meaning. To achieve this, we employ Sentence-BERT (SBERT) \cite{reimers2019sentence}, a transformer-based embedding model that generates high-quality dense vectors at the sentence level. By leveraging the contextual representations learned by BERT, SBERT provides fixed-length embeddings in which semantically similar ideas are mapped to nearby points in vector space. 

These embeddings form the foundation for subsequent analysis. Because SBERT encodes rich contextual information, it enables effective semantic comparisons between ideas. This facilitates the grouping of related sentences and improves the accuracy of topic discovery. By incorporating SBERT, our framework ensures that clusters are based on true semantic similarity rather than surface-level lexical overlap, resulting in more coherent and interpretable topics.

\subsection{Dimension Reduction}

Although SBERT produces embeddings that capture nuanced semantic information, the resulting vectors are high-dimensional. High dimensionality can be problematic: as dimensionality increases, distances between data points tend to become less discriminative, a phenomenon sometimes referred to as the ``curse of dimensionality'' \cite{aggarwal2001surprising}. This effect complicates the performance of clustering algorithms, which rely on meaningful distance metrics to separate related from unrelated data \cite{pandove2018systematic}. Dimensionality reduction provides a practical solution to this challenge by projecting embeddings into a lower-dimensional space while preserving their semantic structure \cite{grootendorst2022bertopic}.

Several well-known techniques exist for this purpose, including Principal Component Analysis (PCA), t-distributed Stochastic Neighbor Embedding (t-SNE), and Uniform Manifold Approximation and Projection (UMAP) \cite{allaoui2020considerably}. In our model, we adopt UMAP because it simultaneously preserves local and global relationships among data points, ensuring that semantically similar ideas remain close together in the reduced space \cite{mcinnes2018umap}. UMAP is also efficient with large datasets and can capture non-linear structures, making it particularly suitable for clustering textual data. By employing UMAP, we achieve more meaningful clusters, which significantly improves the quality of subsequent topic extraction.

\subsection{Document Clustering}

Clustering is the central step of our framework, as it transforms reduced embeddings into semantically coherent groups of ideas. Each resulting cluster is treated as a potential topic, reflecting a subset of brainstorming contributions that share a common theme. This step is critical for evaluating brainstorming performance, since the distribution and coherence of clusters reveal how participants explored and developed different ideas. 

To perform clustering, we use Hierarchical Density-Based Spatial Clustering of Applications with Noise (HDBSCAN), chosen for its ability to discover clusters of varying shapes and densities while handling noisy data. HDBSCAN not only identifies coherent clusters but also labels low-density points as outliers, thereby filtering out irrelevant or weakly related ideas. In addition, it provides hierarchical insights by revealing potential subtopics within broader themes. These properties make HDBSCAN especially effective for modeling brainstorming data, where ideas can range from well-developed to loosely connected fragments. Its robustness and scalability allow us to capture both the structure and the diversity of group discussions in a principled manner.

\subsection{Topic Extraction} 
 Numerous topic modeling studies have demonstrated that the documents within each cluster exhibit a clear association with a specific topic \cite{grootendorst2022bertopic}. In other words, each cluster represents a single topic. However, it is essential to realize that the documents within a cluster may contain multiple topics and subtopics, indicating a certain level of topic diversity within clusters in an absolute sense. 

 In our approach, after applying the HDBSCAN clustering algorithm and identifying clusters, the next step involves building an independent vocabulary and detecting topic words for each cluster, which consists of several steps. Firstly, sentences within each cluster are split into individual words. These words are then mapped to their corresponding contextual embedding values, or word embeddings are generated for each vocabulary word after constructing the vocabulary. Secondly, the average semantic similarity of each topic representative word within the cluster is computed by comparing it with the semantic information of each sentence. 
 Let's describe mathematically as follows:\\
  A cluster consists of a collection of $n$ topic representative words, represented as vocabulary $W$, accompanied by a set of $N$ contextually similar sentences denoted as $S$. Here, $W$ is defined as $W=[w_1, w_2, …, w_n]$, and $S$ is represented as $ S=[s_1, s_2, …, s_N]$, then the average cosine similarity of each word within a cluster defined as:\\
\[ Ave\_cos\_sim (w_i) =  \frac{\sum_{j=1}^{N} cos(\vec{ w_i}, \vec {s_j})} {N} \hspace{1.2cm}  (1)\] where $w_i$ represents a vector associated with a topic word from the vocabulary $W$, encapsulating the semantic information of that word, and $s_j$ denotes a vector associated with a sentence from the set of contextually similar sentences $S$, capturing the semantic information of the entire sentence.

This process provides an average of representative semantic similarity values for each topic word in that cluster. Subsequently, candidate topic words are organized and sorted based on the average semantic similarity values, and the top 'k' words are selected from each cluster. This process enables the extraction of topics from each cluster with enhanced accuracy and relevance of topic words specific to that cluster. Finally, topic refinement is performed; after extracting the topics, it is essential to consider the dissimilarity between topics. Hence, we merge the cluster-based least common topic representation with its most similar counterparts through an iterative process using similarity measurement techniques. This iterative process helps us reduce the number of topics to a user-specified value. This iterative process helps us to reduce the number of topics to a user-specified value.

\section{Experiments and Results}
\subsection{Data Collection}
The data was collected through collaborative brainstorming sessions involving multiple groups in virtual meetings, including three to five participants in each group. The brainstorming sessions were conducted using Zoom videoconferencing modality. Each group participated in a 60-minute brainstorming session. The groups were comprised of graduate and undergraduate students representing various majors in a US university. During the brainstorming sessions, the groups were instructed to generate ideas on ways to improve their university. As a result of these virtual meetings, a large amount of transcript/text data was extracted and collected, containing the wealth of creative thinking and insights offered by each participant in each group. We conducted the following modular-based tasks to process these large amounts of generated text data to gain important insights.

\subsection{Sentence Embedding}
Sentence embedding is the initial step toward measuring contextual similarity between each participant's generated idea/sentence. The sentence embedding process enables us to understand the contextual nuances of each idea and the relationships between one idea with another, leading to more comprehensive understandings and valuable information for the next steps. We feed the entire documents (collection of ideas) into the SBERT model to encode them into dense and fixed-length numerical vectors called sentence embedding, which is a high-dimensional vector. These numerical vector representations of the sentence(idea) are used to group the ideas based on their semantic similarity.

\subsection{Dimension Reduction}
The SBERT model generates high-dimensional vectors/sentence embeddings. These high-dimensional vector outputs are suitable for dealing with complex semantic relationships between the generated ideas. However, these high-dimensional vector representations can have computational challenges for some machine learning algorithms. To mitigate these problems, we employed the UMAP (Uniform Manifold Approximation and Projection) dimension reduction technique to transform these high-dimensional vector representations into 2-dimensional vector spaces.  The following figure illustrates the complete set of generated ideas or sentences represented through sentence embedding and dimension reduction methods.

\begin{figure}[h] 
\centering
{\includegraphics [width=0.4\textwidth]{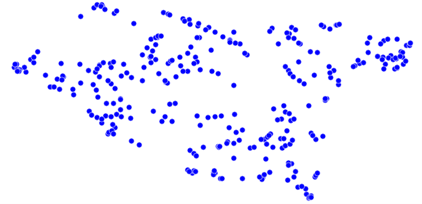}}
\caption{Exploring Individual Ideas/Sentences: Unclustered and Diverse Perspectives}
\label{fig:unclusterDataset}
\end{figure}

\subsection{Clustering Ideas/Sentences}
\begin{figure}[h] 
\centering
{\includegraphics [width=0.4\textwidth]{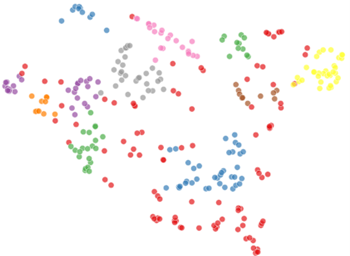}}
\caption{Clustered Ideas/Sentences Based on Contextual Similarity: Discovering Groupings and Outliers}
\label{fig:clusterWithOutliers}
\end{figure}

The ideas are represented numerically in the embedding module and, consequently, reduced into lower-dimension vector spaces for more efficient clustering purposes. The employed HDBSCAN algorithm grouped these lower-dimension vector space representations of ideas into distinct clusters based on their semantic similarity and identified unrelated ideas as outliers. Now, we have a number of clusters that contain semantically similar ideas and outliers.  Figure ~\ref{fig:clusterWithOutliers} shows the clustered ideas and outlier ideas. 

\textbf{Handling Outliers:} Outliers are ideas/sentences that lack strong relationships with each other and with other ideas/sentences. The HDBSCAN algorithm not only identifies contextually similar sentences/ideas but also detects contextually unrelated sentences/ideas, labeling them as outliers or noises (denoted by -1).  We do not include these outlier sentences/ideas in our topic modeling process because they exhibit weaker relationships among themselves and with other distinct clusters. As depicted in the figure ~\ref{fig:clusterWithOutOutliers} below, the outlier ideas are excluded, and the distinct clusters are preserved for the topic extraction process. The forthcoming steps comprise preprocessing each sentence /idea within the clusters and extracting the hidden topics from each cluster.

\begin{figure}[h] 
\centering
{\includegraphics [width=0.4\textwidth]{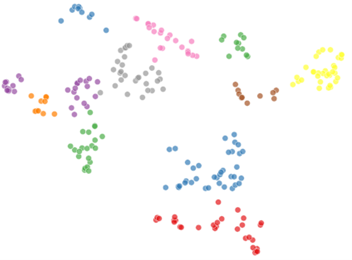}}
\caption{Clustered Ideas/Sentences Based on Contextual Similarity: Excluding Outliers}
\label{fig:clusterWithOutOutliers}
\end{figure}

In the previous steps, we processed distinct clusters that contained contextually similar ideas. However, the representation of each cluster remains unidentified or hidden. Therefore, extracting latent topics from these distinct clusters becomes imperative to explore the information and insights each cluster encapsulates. Hence, some fundamental further steps and tasks, such as preprocessing and topic extraction, are also required to achieve this.

\captionsetup[table]{skip=20pt} 
\begin{table*}[hbt!]
\small
\centering 
{\begin{tabular}{ p{0.046\textwidth} p{0.067\textwidth} p{10.5cm} c} 
\hline
\textbf{Cluster}&\textbf{Words/ Cluster} & \textbf{\   \  List of Topic Words/Ideas of Dataset1 } & \textbf{TC} \\\hline
1 &134&education, academic, curriculum, degree, job, student, psych, benefit, study, worker  & 0.736 \\
2 &52 &bookstore, book, vending, store, printing, merchandise, money, spend, buy, purchase, print  & 0.714\\
3 & 46&scholarship, tuition, aid, college, financial, academic, student, degree, providing, apply  & 0.733\\
4 &119 &parking, ticketing, driving, garage, building, field, enforce, outside, commuter, ticket  & 0.609\\
5 &126 &dining, lunch, menu, food,  hall, room, campus, seating, providing, hour  & 0.643\\
6 &92 &fraternity, sorority, social, diversity,  collaboration, inclusion, diverse, School, hazing, campus   & 0.673\\
7 & 56&class, lab, improve,  course, science,  semester, testing, subject,  session, chemistry  &0.488 \\
8 &55 &housing, security, architecture, residence, maintenance, access, crime,  elevator, improve, fix  & 0.608\\
9 &26 &schedule, week, day, weekend, hour, class, change, ahead  & 0.464\\
10 &66 &advertising, communication, event, notification, information, campus, advance  &0.506 \\

&&\textbf{Overall Topic Coherence} &\textbf{0.617}
 \\
 \hline
\end{tabular}} 

\caption{Topics, top 10 words, and c\_v individual topic coherence scores with overall topic coherence score as the average of individual score from a single program run output.}

\label{Table:topic results}
\end{table*}

\subsubsection{Pre-processing}
Each distinctly identified cluster contains a collection of ideas that are contextually similar in meaning. These sentences(ideas) might comprise common stop words (such as `of', `a,' `the', and `is'), punctuation marks, non-English words (like `sfdeg' `s23d2', and `21hy\%), and special characters that are not considered meaningful as topic words. Topic words are mostly nominated from verbs and nouns, even though the selection may vary based on the type of problem and the nature of the input text data. Eliminating unimportant words, punction marks, and characters is very important to enhance the quality of our topics. The following two fundamental tasks are required to achieve this.

\subsubsection{Vocabulary and Word Embeddings}
After completing the preprocessing task, we create an independent vocabulary for each cluster and generate word embeddings for each vocabulary word. Next, we maintain all these word-embedding pair collections separately for each cluster based on the designated cluster index. In this way, we organize a comprehensive collection of word-embedding associations for each cluster.

\begin{figure}[h] 
\centering
{\includegraphics [width=0.45\textwidth]{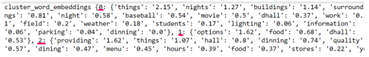}}
\caption{A cluster-based vocabulary contains word embeddings associated with topic words}
\label{fig:vocabulary}
\end{figure}

\subsubsection{Topic Extraction}
Numerous topic modeling studies have demonstrated that the documents within each cluster exhibit a clear association with a specific topic \cite{grootendorst2022bertopic}.
In the previous steps, we effectively extracted each cluster's most relevant representative words, along with their corresponding embeddings. These topic words were computed using Equation 1. Subsequently, we sorted the candidate topic words in descending order based on the average similarity values within each cluster. Finally, we selected the top k words that most accurately represent each cluster. The results are presented in Table ~\ref{Table:topic results}.

\subsubsection{Topic Refinement}
Topic refinement is performed after the topic extraction process is completed to reduce the number of topics to a user-specified value. If a topic is detected in a cluster and exhibits a low cosine similarity value with other topics, it is preserved as a unique topic. We merge the least common topic with its most similar counterparts through an iterative process using similarity measure techniques such as cosine similarity.

\subsubsection{Automatic Model Evaluation and Comparison} Model evaluation is the most crucial task \cite{mersha2025evaluating, mersha2025unified}.
The model's performance is evaluated using the Gensim topic coherence metric, a well-known Python open-source library, a widely accepted and easily applicable metric for assessing the quality of topics. More precisely, we used the CV Coherence metric to measure the coherence of the generated topics, enabling us to understand and interpret the meaning of each topic \cite{roder2015exploring}. The CV coherence metric has comparatively approximated human judgment \cite{lau2014machine}; its score ranges between -1 and 1, 1 indicating a perfect association of topic words. These individual topic scores indicate that the top k words in each topic have a stronger relationship and a high probability of co-occurring together and have semantic similarity within the context of the given clustered ideas. The overall coherence score of the model is computed by averaging those individual topic coherence scores as shown in Table ~\ref{Table:topic results}.

We conducted an extensive performance comparison between our proposed model and well-known, established models, including  Latent Dirichlet Allocation (LDA),   Topic Modeling in Embedding Spaces (ETM), and BERTopic.  Specifically, LDA is employed with a probabilistic method,  ETM is implemented in an embedding-based approach, and BERTopic is a TF\_IDF and embedding-based technique. Topic Coherence (TC) is computed for each topic model, varying the number of topics from 2 to 10 with increments of 2. We averaged the outputs from three separate runs at each interval to enhance consistency, resulting in an average score derived from a cumulative total of 15 distinct runs. Table~\ref{tbl:comparison} shows the model comparison results .\\

\begin{table}[t] 
\normalsize
\centering
\renewcommand{\arraystretch}{1.5}
\begin{tabular}{ p{2cm}|p{2.1cm}|p{2cm} }
 \hline
 \multicolumn{3}{c} {\textbf{Model Comparison Results for Dataset1} } \\
 \hline
 
\textbf{ Models}& \textbf{TC(C\_V)} & \textbf{TC(C\_npmi)}\\
 \hline
 LDA  &0.462    &0.066\\

 ETM &0.544		&0.094\\

 BERTopic &0.552		&0.192\\

 Our Model & \colorbox{green}{0.687}		& \colorbox{green}{0.264}\\
 
 \hline

\end{tabular}
\caption{ Model comparison with C\_V and C\_npmi metrics results}
\label{tbl:comparison}
\end{table} 


During meetings, a substantial number of ideas are generated, making it challenging for humans to effectively organize and discover underlying themes and patterns in these ideas. As we discussed in the previous sections, this paper aims to cluster ideas generated by brainstorm session teams based on contextual similarity of the ideas and then extract the most representative topic for each cluster of such contextually similar ideas. Each cluster is a mixture of topics that facilitates the interpretation of the hidden themes within the clusters. We have extracted these topics and ranked them in descending order to identify the most representative topics for each cluster, as shown in Table ~\ref{Table:topic results}. In the representative list of words for each cluster, the first word, such as "education" in cluster 1, exhibits the highest semantic similarity to all the ideas within the respective cluster. Hence, we can extract the most significant insights from the meeting using these topics. In this case, we can identify the top 10 general topics in the meeting session, which categorize all the generated ideas. Furthermore, we can analyze the relationships between topics in a human context. For example, topic 9, 'schedule,' has some defined relations like activity to topic 7, 'class,' and topic 5, 'dining.' 

\section{Future Work}
In future work, we plan to integrate Explainable AI (XAI) techniques to enhance the interpretability of our framework. While the current approach effectively clusters and extracts topics, XAI methods could reveal which words and sentences most strongly influence cluster assignments and topic coherence \cite{mersha2024explainable,mersha2024explainability,mersha2025explainable, mersha2026explainable}. This added transparency would not only validate the quality of the discovered topics but also provide deeper insights into how divergent and convergent thinking emerge in group discussions.

\section{Conclusion}
This study presented a semantic-driven topic modeling framework designed to analyze ideas generated during virtual brainstorming sessions. By integrating Sentence-BERT embeddings, UMAP dimensionality reduction, HDBSCAN clustering, and topic extraction with refinement, the framework effectively organized large volumes of textual data into coherent and interpretable themes. Applied to structured Zoom meetings, the approach not only filtered out noise and identified outliers but also revealed meaningful topics that captured both convergent and divergent dimensions of group creativity.

Our experimental results demonstrated that the proposed model achieved higher topic coherence than established baseline methods, confirming the advantage of embedding-based methods in capturing contextual semantic relationships. The ability to automatically categorize ideas and evaluate the depth of topic exploration provides valuable insights into group dynamics, supporting the study of creativity in collaborative contexts.

Overall, this work contributes a scalable and interpretable framework for studying creativity in synchronous virtual collaboration. Future research may extend this approach by incorporating multimodal signals (e.g., audio and video cues), exploring cross-domain applications, and integrating large language models to facilitate brainstorming sessions in real-time.

\bibliographystyle{elsarticle-num}
\bibliography{bib.bib}

@article{mersha2024semantic,
  title={Semantic-driven topic modeling using transformer-based embeddings and clustering algorithms},
  author={Mersha, Melkamu Abay and Kalita, Jugal and others},
  journal={Procedia Computer Science},
  volume={244},
  pages={121--132},
  year={2024},
  publisher={Elsevier}
}

@article{mersha2026explainable,
  title={Explainable AI: Context-aware layer-wise integrated gradients for explaining transformer models},
  author={Mersha, Melkamu Abay and Kalita, Jugal},
  journal={Neurocomputing},
  pages={133050},
  year={2026},
  publisher={Elsevier}
}

@article{rodriguez2020computational,
  title={A computational social science perspective on qualitative data exploration: Using topic models for the descriptive analysis of social media data},
  author={Rodriguez, Maria Y and Storer, Heather},
  journal={Journal of Technology in Human Services},
  volume={38},
  number={1},
  pages={54--86},
  year={2020},
  publisher={Taylor \& Francis}
}

@article{joshi2023deepsumm,
  title={DeepSumm: Exploiting topic models and sequence to sequence networks for extractive text summarization},
  author={Joshi, Akanksha and Fidalgo, Eduardo and Alegre, Enrique and Fern{\'a}ndez-Robles, Laura},
  journal={Expert Systems with Applications},
  volume={211},
  pages={118442},
  year={2023},
  publisher={Elsevier}
}

@article{albalawi2020using,
  title={Using topic modeling methods for short-text data: A comparative analysis},
  author={Albalawi, Rania and Yeap, Tet Hin and Benyoucef, Morad},
  journal={Frontiers in artificial intelligence},
  volume={3},
  pages={42},
  year={2020},
  publisher={Frontiers Media SA}
}

@article{mustak2021artificial,
  title={Artificial intelligence in marketing: Topic modeling, scientometric analysis, and research agenda},
  author={Mustak, Mekhail and Salminen, Joni and Pl{\'e}, Lo{\"\i}c and Wirtz, Jochen},
  journal={Journal of Business Research},
  volume={124},
  pages={389--404},
  year={2021},
  publisher={Elsevier}
}

@article{sbalchiero2020topic,
  title={Topic modeling, long texts and the best number of topics. Some Problems and solutions},
  author={Sbalchiero, Stefano and Eder, Maciej},
  journal={Quality \& Quantity},
  volume={54},
  pages={1095--1108},
  year={2020},
  publisher={Springer}
}

@article{punziano2023digital,
  title={Digital mixed content analysis for the study of digital platform social data: An illustration from the analysis of COVID-19 risk perception in the Italian twittersphere},
  author={Punziano, Gabriella and De Falco, Ciro C and Trezza, Domenico},
  journal={Journal of Mixed Methods Research},
  volume={17},
  number={2},
  pages={143--170},
  year={2023},
  publisher={SAGE Publications Sage CA: Los Angeles, CA}
}

@article{blei2003latent,
  title={Latent dirichlet allocation},
  author={Blei, David M and Ng, Andrew Y and Jordan, Michael I},
  journal={Journal of machine Learning research},
  volume={3},
  number={Jan},
  pages={993--1022},
  year={2003}
}

@article{fevotte2011algorithms,
  title={Algorithms for nonnegative matrix factorization with the $\beta$-divergence},
  author={F{\'e}votte, C{\'e}dric and Idier, J{\'e}r{\^o}me},
  journal={Neural computation},
  volume={23},
  number={9},
  pages={2421--2456},
  year={2011},
  publisher={MIT Press}
}

@inproceedings{hofmann1999probabilistic,
  title={Probabilistic latent semantic indexing},
  author={Hofmann, Thomas},
  booktitle={Proceedings of the 22nd annual international ACM SIGIR conference on Research and development in information retrieval},
  pages={50--57},
  year={1999}
}

@article{grootendorst2022bertopic,
  title={BERTopic: Neural topic modeling with a class-based TF-IDF procedure},
  author={Grootendorst, Maarten},
  journal={arXiv preprint arXiv:2203.05794},
  year={2022}
}

@article{reimers2019sentence,
  title={Sentence-bert: Sentence embeddings using siamese bert-networks},
  author={Reimers, Nils and Gurevych, Iryna},
  journal={arXiv preprint arXiv:1908.10084},
  year={2019}
}

@inproceedings{aggarwal2001surprising,
  title={On the surprising behavior of distance metrics in high dimensional space},
  author={Aggarwal, Charu C and Hinneburg, Alexander and Keim, Daniel A},
  booktitle={Database Theory—ICDT 2001: 8th International Conference London, UK, January 4--6, 2001 Proceedings 8},
  pages={420--434},
  year={2001},
  organization={Springer}
}

@article{pandove2018systematic,
  title={Systematic review of clustering high-dimensional and large datasets},
  author={Pandove, Divya and Goel, Shivan and Rani, Rinkl},
  journal={ACM Transactions on Knowledge Discovery from Data (TKDD)},
  volume={12},
  number={2},
  pages={1--68},
  year={2018},
  publisher={ACM New York, NY, USA}
}

@article{mcinnes2018umap,
  title={Umap: Uniform manifold approximation and projection for dimension reduction},
  author={McInnes, Leland and Healy, John and Melville, James},
  journal={arXiv preprint arXiv:1802.03426},
  year={2018}
}

@inproceedings{allaoui2020considerably,
  title={Considerably improving clustering algorithms using UMAP dimensionality reduction technique: A comparative study},
  author={Allaoui, Mebarka and Kherfi, Mohammed Lamine and Cheriet, Abdelhakim},
  booktitle={International conference on image and signal processing},
  pages={317--325},
  year={2020},
  organization={Springer}
}

@article{bianchi2020cross,
  title={Cross-lingual contextualized topic models with zero-shot learning},
  author={Bianchi, Federico and Terragni, Silvia and Hovy, Dirk and Nozza, Debora and Fersini, Elisabetta},
  journal={arXiv preprint arXiv:2004.07737},
  year={2020}
}

@article{blei2012probabilistic,
  title={Probabilistic topic models},
  author={Blei, David M},
  journal={Communications of the ACM},
  volume={55},
  number={4},
  pages={77--84},
  year={2012},
  publisher={ACM New York, NY, USA}
}

@inproceedings{roder2015exploring,
  title={Exploring the space of topic coherence measures},
  author={R{\"o}der, Michael and Both, Andreas and Hinneburg, Alexander},
  booktitle={Proceedings of the eighth ACM international conference on Web search and data mining},
  pages={399--408},
  year={2015}
}

@inproceedings{lau2014machine,
  title={Machine reading tea leaves: Automatically evaluating topic coherence and topic model quality},
  author={Lau, Jey Han and Newman, David and Baldwin, Timothy},
  booktitle={Proceedings of the 14th Conference of the European Chapter of the Association for Computational Linguistics},
  pages={530--539},
  year={2014}
}

@article{kim2019insider,
  title={Insider threat detection based on user behavior modeling and anomaly detection algorithms},
  author={Kim, Junhong and Park, Minsik and Kim, Haedong and Cho, Suhyoun and Kang, Pilsung},
  journal={Applied Sciences},
  volume={9},
  number={19},
  pages={4018},
  year={2019},
  publisher={MDPI}
}

@article{uthirapathy2023topic,
  title={Topic Modelling and Opinion Analysis On Climate Change Twitter Data Using LDA And BERT Model.},
  author={Uthirapathy, Samson Ebenezar and Sandanam, Domnic},
  journal={Procedia Computer Science},
  volume={218},
  pages={908--917},
  year={2023},
  publisher={Elsevier}
}

@article{koh2021topic,
  title={Topic modeling as a tool for analyzing library chat transcripts},
  author={Koh, HyunSeung and Fienup, Mark},
  journal={Information Technology and Libraries},
  volume={40},
  number={3},
  year={2021}
}

@article{chin2023user,
  title={User-chatbot conversations during the COVID-19 pandemic: study based on topic modeling and sentiment analysis},
  author={Chin, Hyojin and Lima, Gabriel and Shin, Mingi and Zhunis, Assem and Cha, Chiyoung and Choi, Junghoi and Cha, Meeyoung},
  journal={Journal of medical Internet research},
  volume={25},
  pages={e40922},
  year={2023},
  publisher={JMIR Publications Toronto, Canada}
}

@inproceedings{kim2008meeting,
  title={Meeting mediator: enhancing group collaborationusing sociometric feedback},
  author={Kim, Taemie and Chang, Agnes and Holland, Lindsey and Pentland, Alex Sandy},
  booktitle={Proceedings of the 2008 ACM conference on Computer supported cooperative work},
  pages={457--466},
  year={2008}
}

@article{nijstad1999persistence,
  title={Persistence of brainstorming groups: How do people know when to stop?},
  author={Nijstad, Bernard A and Stroebe, Wolfgang and Lodewijkx, Hein FM},
  journal={Journal of Experimental Social Psychology},
  volume={35},
  number={2},
  pages={165--185},
  year={1999},
  publisher={Elsevier}
}

@article{nijstad2002cognitive,
  title={Cognitive stimulation and interference in groups: Exposure effects in an idea generation task},
  author={Nijstad, Bernard A and Stroebe, Wolfgang and Lodewijkx, Hein FM},
  journal={Journal of experimental social psychology},
  volume={38},
  number={6},
  pages={535--544},
  year={2002},
  publisher={Elsevier}
}

@article{huber2019automatically,
  title={Automatically analyzing brainstorming language behavior with Meeter},
  author={Huber, Bernd and Shieber, Stuart and Gajos, Krzysztof Z},
  journal={Proceedings of the ACM on human-computer interaction},
  volume={3},
  number={CSCW},
  pages={1--17},
  year={2019},
  publisher={ACM New York, NY, USA}
}

@article{rietzschel2007relative,
  title={Relative accessibility of domain knowledge and creativity: The effects of knowledge activation on the quantity and originality of generated ideas},
  author={Rietzschel, Eric F and Nijstad, Bernard A and Stroebe, Wolfgang},
  journal={Journal of experimental social psychology},
  volume={43},
  number={6},
  pages={933--946},
  year={2007},
  publisher={Elsevier}
}

@article{ferguson2022communication,
  title={Communication patterns in engineering enterprise social networks: an exploratory analysis using short text topic modelling},
  author={Ferguson, Sharon A and Cheng, Kathy and Adolphe, Lauren and Van de Zande, Georgia and Wallace, David and Olechowski, Alison},
  journal={Design Science},
  volume={8},
  pages={e18},
  year={2022},
  publisher={Cambridge University Press}
}

@article{zhao2021topic,
  title={Topic modelling meets deep neural networks: A survey},
  author={Zhao, He and Phung, Dinh and Huynh, Viet and Jin, Yuan and Du, Lan and Buntine, Wray},
  journal={arXiv preprint arXiv:2103.00498},
  year={2021}
}

@inproceedings{terragni2021octis,
  title={OCTIS: Comparing and optimizing topic models is simple!},
  author={Terragni, Silvia and Fersini, Elisabetta and Galuzzi, Bruno Giovanni and Tropeano, Pietro and Candelieri, Antonio},
  booktitle={Proceedings of the 16th Conference of the European Chapter of the Association for Computational Linguistics: System Demonstrations},
  pages={263--270},
  year={2021}
}

@inproceedings{agarwal2021comparative,
  title={Comparative Study of Topic Modeling and Word Embedding Approaches for Web Service Clustering},
  author={Agarwal, Neha and Sikka, Geeta and Awasthi, Lalit Kumar},
  booktitle={2021 Thirteenth International Conference on Contemporary Computing (IC3-2021)},
  pages={309--313},
  year={2021}
}

@inproceedings{qiang2017topic,
  title={Topic modeling over short texts by incorporating word embeddings},
  author={Qiang, Jipeng and Chen, Ping and Wang, Tong and Wu, Xindong},
  booktitle={Advances in Knowledge Discovery and Data Mining: 21st Pacific-Asia Conference, PAKDD 2017, Jeju, South Korea, May 23-26, 2017, Proceedings, Part II 21},
  pages={363--374},
  year={2017},
  organization={Springer}
}

@article{dieng2020topic,
  title={Topic modeling in embedding spaces},
  author={Dieng, Adji B and Ruiz, Francisco JR and Blei, David M},
  journal={Transactions of the Association for Computational Linguistics},
  volume={8},
  pages={439--453},
  year={2020},
  publisher={MIT Press One Rogers Street, Cambridge, MA 02142-1209, USA journals-info~…}
}

@inproceedings{tang2023research,
  title={Research on the Evolution of Journal Topic Mining Based on the BERT-LDA Model},
  author={Tang, Guofeng and Chen, Xuhui and Li, Ning and Cui, Jianfeng},
  booktitle={SHS Web of Conferences},
  volume={152},
  pages={03012},
  year={2023},
  organization={EDP Sciences}
}

@article{alshami2024smart,
  title={Smart-vision: survey of modern action recognition techniques in vision},
  author={AlShami, Ali K and Rabinowitz, Ryan and Lam, Khang and Shleibik, Yousra and Mersha, Melkamu and Boult, Terrance and Kalita, Jugal},
  journal={Multimedia Tools and Applications},
  pages={1--72},
  year={2024},
  publisher={Springer}
}

@inproceedings{trenk2024text,
  title={Text summarization using rhetorical structure trees},
  author={Trenk, Edward and Mersha, Melkamu and Kalita, Jugal},
  booktitle={Recent Advances in Natural Language Processing},
  pages={72},
  year={2024}
}

@article{mersha2025explainable,
  title={Explainable AI: XAI-guided context-aware data augmentation},
  author={Mersha, Melkamu Abay and Yigezu, Mesay Gemeda and Tonja, Atnafu Lambebo and Shakil, Hassan and Iskander, Samer and Kolesnikova, Olga and Kalita, Jugal},
  journal={Expert Systems with Applications},
  volume={289},
  pages={128364},
  year={2025},
  publisher={Elsevier}
}

@article{tonja2023first,
  title={First Attempt at Building Parallel Corpora for Machine Translation of Northeast India's Very Low-Resource Languages},
  author={Tonja, Atnafu Lambebo and Mersha, Melkamu and Kalita, Ananya and Kolesnikova, Olga and Kalita, Jugal},
  journal={arXiv preprint arXiv:2312.04764},
  year={2023}
}

@article{mersha2025unified,
  title={A unified framework with novel metrics for evaluating the effectiveness of xai techniques in llms},
  author={Mersha, Melkamu Abay and Yigezu, Mesay Gemeda and Shakil, Hassan and AlShami, Ali K and Byun, Sanghyun and Kalita, Jugal},
  journal={arXiv preprint arXiv:2503.05050},
  year={2025}
}

@article{mersha2024explainability,
  title={Explainability in neural networks for natural language processing tasks},
  author={Mersha, Melkamu and Bitewa, Mingiziem and Abay, Tsion and Kalita, Jugal},
  journal={arXiv preprint arXiv:2412.18036},
  year={2024}
}

@article{mersha2024ethio,
  title={Ethio-fake: Cutting-edge approaches to combat fake news in under-resourced languages using explainable ai},
  author={Mersha, Melkamu Abay and Bade, Girma Yohannis and Kalita, Jugal and Kolesnikova, Olga and Gelbukh, Alexander and others},
  journal={Procedia Computer Science},
  volume={244},
  pages={133--142},
  year={2024},
  publisher={Elsevier}
}

@article{yigezu2024ethio,
  title={Ethio-fake: Cutting-edge approaches to combat fake news in under-resourced languages using explainable ai},
  author={Yigezu, Mesay Gemeda and Mersha, Melkamu Abay and Bade, Girma Yohannis and Kalita, Jugal and Kolesnikova, Olga and Gelbukh, Alexander},
  journal={arXiv preprint arXiv:2410.02609},
  year={2024}
}

@article{mersha2025evaluating,
  title={Evaluating the effectiveness of XAI techniques for encoder-based language models},
  author={Mersha, Melkamu Abay and Yigezu, Mesay Gemeda and Kalita, Jugal},
  journal={Knowledge-Based Systems},
  volume={310},
  pages={113042},
  year={2025},
  publisher={Elsevier}
}

@article{mersha2024explainable,
  title={Explainable artificial intelligence: A survey of needs, techniques, applications, and future direction},
  author={Mersha, Melkamu and Lam, Khang and Wood, Joseph and Alshami, Ali K and Kalita, Jugal},
  journal={Neurocomputing},
  volume={599},
  pages={128111},
  year={2024},
  publisher={Elsevier}
}

@article{nijstad2006group,
  title={How the group affects the mind: A cognitive model of idea generation in groups},
  author={Nijstad, Bernard A and Stroebe, Wolfgang},
  journal={Personality and social psychology review},
  volume={10},
  number={3},
  pages={186--213},
  year={2006},
  publisher={Sage Publications Sage CA: Los Angeles, CA}
}

@article{ziegler2000idea,
  title={Idea production in nominal and virtual groups: Does computer-mediated communication improve group brainstorming?},
  author={Ziegler, Rene and Diehl, Michael and Zijlstra, Gavin},
  journal={Group Processes \& Intergroup Relations},
  volume={3},
  number={2},
  pages={141--158},
  year={2000},
  publisher={Sage Publications Sage CA: Thousand Oaks, CA}
}

@article{rietzschel2010selection,
  title={The selection of creative ideas after individual idea generation: Choosing between creativity and impact},
  author={Rietzschel, Eric F and Nijstad, Bernard A and Stroebe, Wolfgang},
  journal={British journal of psychology},
  volume={101},
  number={1},
  pages={47--68},
  year={2010},
  publisher={Wiley Online Library}
}

@article{paulus2007toward,
  title={Toward more creative and innovative group idea generation: A cognitive-social-motivational perspective of brainstorming},
  author={Paulus, Paul B and Brown, Vincent R},
  journal={Social and Personality Psychology Compass},
  volume={1},
  number={1},
  pages={248--265},
  year={2007},
  publisher={Wiley Online Library}
}

\end {document}